\documentclass[sigconf,authorversion,nonacm]{acmart}

\usepackage{booktabs} 
\usepackage{subfig}
\usepackage{graphicx}
\usepackage{colortbl}
\usepackage{algorithm}
\usepackage[noend]{algorithmic}
\usepackage{bm}
\usepackage{multirow}
\usepackage{verbatim}
\usepackage[normalem]{ulem}
\usepackage{xcolor}
\usepackage{enumitem}
\setlist{leftmargin=4mm}

\usepackage{mathtools}
\DeclarePairedDelimiter\ceil{\lceil}{\rceil}

\usepackage[T1]{fontenc}
\usepackage{lmodern}
\usepackage{natbib}
\newcounter{ALC@tempcntr}

\definecolor{mypink}{cmyk}{0, 0.7808, 0.4429, 0.1412}

\definecolor{blue-light}{RGB}{66, 191, 244}

\newcommand{\ignore}[1]{}

\setcopyright{none}

\acmConference[]{}
\acmBooktitle{}
\acmPrice{}
\acmDOI{}
\acmISBN{}

\begin{document}


\sloppy 







\title{On the Fitness Landscapes of Interdependency Models in the Travelling Thief Problem}

\renewcommand{\shorttitle}{On the Fitness Landscapes of Interdependency Models in the Travelling Thief Problem}

\author{Mohamed El Yafrani}
\affiliation{
\institution{Aalborg University}
\country{Denmark}
}
\email{mey@mp.aau.dk}

\author{Marcella Scoczynski}
\affiliation{
\institution{Federal University of Technology Paran\'a (UTFPR)}
\country{Brazil}
}
\email{marcella@utfpr.edu.br}

\author{Myriam R. B. S. Delgado}
\affiliation{
 \institution{Federal University of Technology - Paran\'{a} (UTFPR)}
  \country{Brazil}
 }
\email{myriamdelg@utfpr.edu.br}

\author{Ricardo L\"{u}ders}
\affiliation{
 \institution{Federal University of Technology - Paran\'{a} (UTFPR)}
 \country{Brazil}
 }
\email{luders@utfpr.edu.br}

\author{Peter Nielsen}
\affiliation{
 \institution{Aalborg University}
 \country{Denmark}
 }
\email{peter@mp.aau.dk}

\author{Markus Wagner}
\affiliation{
 \institution{The University of Adelaide}
 \country{Australia}
 }
\email{markus.wagner@adelaide.edu.au}

\renewcommand{\shortauthors}{El Yafrani et. al.}

\begin{abstract}
Since its inception in 2013, the Travelling Thief Problem (TTP) has been widely studied as an example of problems with multiple interconnected sub-problems.
The dependency in this model arises when tying the travelling time of the ``thief'' to the weight of the knapsack. However, other forms of dependency as well as combinations of dependencies should be considered for investigation, as they are often found in complex real-world problems. Our goal is to study the impact of different forms of dependency in the TTP using a simple local search algorithm.
To achieve this, we use Local Optima Networks, a technique for analysing the fitness landscape. 

\end{abstract}


\maketitle

\section{Motivation}
\label{sec:sect1}


Many real-world optimisation problems can be modelled as a combination of multiple sub-problems with internal dependencies~\cite{bonyadi2019evolutionary}. These dependencies can be formulated as equations or inequalities to connect the sub-problems. 
This class of problems are referred to as problems with multiple interdependent components~\cite{bonyadi2013travelling} or multi-hard problems~\cite{przybylek2018decomposition}.

Due to the lack of a good benchmarking model to study multi-hard problems, the travelling thief problem was introduced by~\citet{bonyadi2013travelling} as a combination of the Travelling Salesman Problem and the Knapsack Problem. A simplified formulation was then introduced by \citet{polyakovskiy2014comprehensive} in order to have a version that is more focused on the interdependency aspect. Since then, several papers introduced solution methods (e.g., \cite{mei2014improving,yafrani2016population,wagner2016stealing,yafrani2018efficiently}), while only few researchers tried to analyse the problem itself empirically and theoretically (e.g., \cite{wu2016impact,yafrani2018fitness,wuijts2019investigation}). Therefore, the goal of this study is to reduce the gap in the problem analysis and provide insights into the problem search landscape.


The standard TTP formulation in~\cite{polyakovskiy2014comprehensive} considers a combination of the Travelling Salesman Problem (TSP) and the Knapsack Problem(KP). The problem considers a set of items scattered in different cities, and a thief that should visit each city exactly once, stealing items on the way and returning to the starting city, while trying to maximise his gain. The dependency in this formulation is modelled by penalising the travel velocity with the knapsack load. While this is a useful model to reflect the dependencies faced in some realistic problems, other problems can embed other -- potentially more complex -- forms of interdependency, where multiple dependency equations or inequalities can be combined~\cite{bonyadi2019evolutionary}.

Herein, we present preliminary results on $4$ dependency models of the TTP. Specifically, we use Local Optima Networks (LONs)~\cite{ochoa2008study} -- a method for fitness landscape analysis -- to assess the difficulty of the standard TTP model and other three proposed models of dependency, including a dependency-free model. The reported results show initial insights into the impact of adding more interdependency equations and the influence of instance features on the difficulty of solving multi-hard problems with local search heuristics.

\section{Proposed approach}

\subsection{Background on Local Optimal Networks}

In the following, we analyse TTP dependency models based on Local Optima Networks (LONs), which is a fitness landscape technique. LONs provide a compressed and simplified version of the search space, represented as a graph where nodes are the local optima and edges are the possible search transitions among optima depending on a given local search operator~\cite{ochoa2008study}. 
Each local optimum has an associated basin of attraction composed of all solutions that converge to it when applying a local search heuristic.

\begin{figure}[htbp]
\centering
\includegraphics[width=0.3\textwidth]{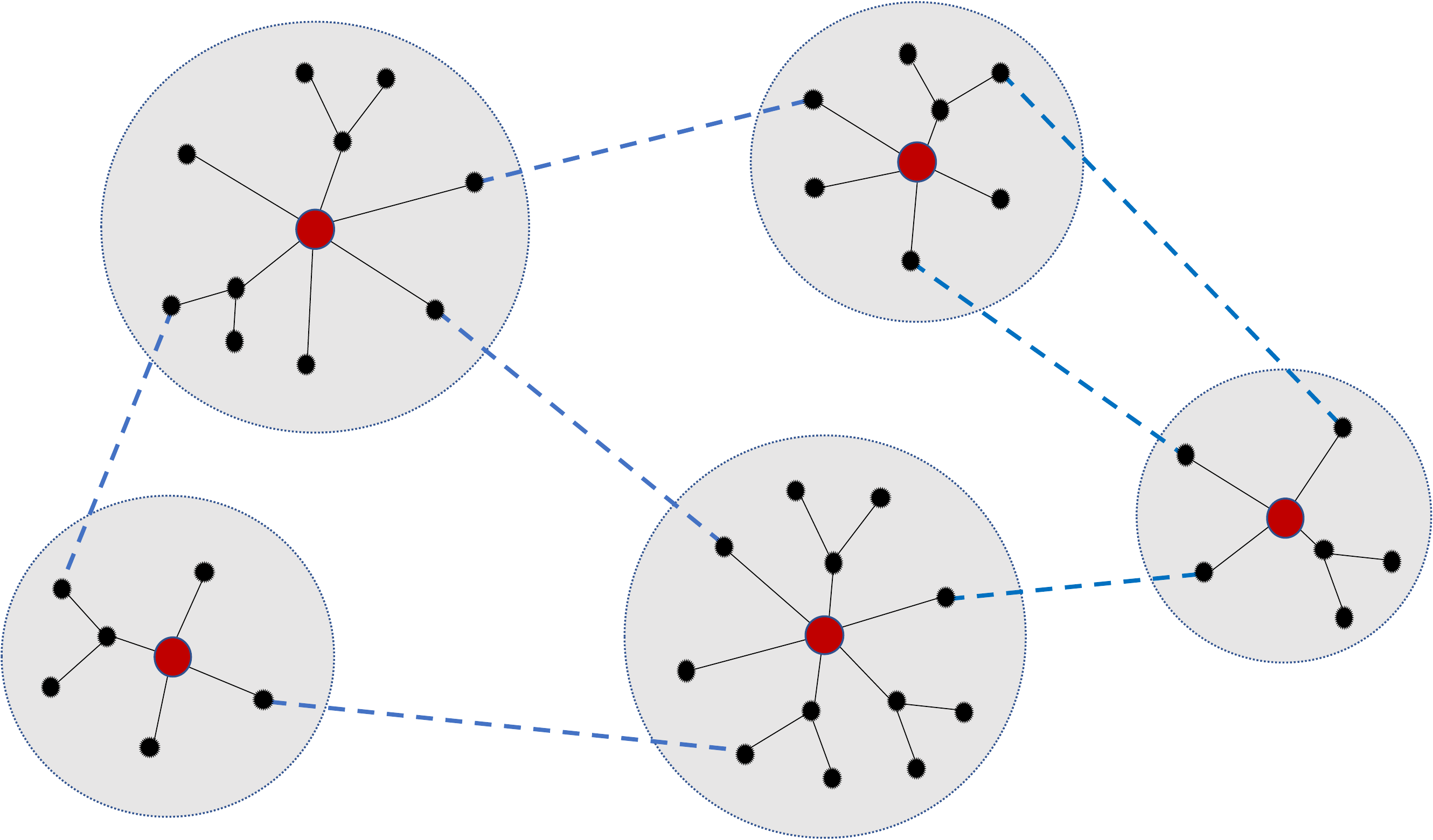}
\caption{An example of a LON with basins of attraction.}
\label{fig:LON-basins}
\end{figure}

As shown in Figure~\ref{fig:LON-basins}, the basin of attraction associated with a local optima $i$ (red dot) is the set $B_{i}=\{s\in S, \mathcal{A}(s)=i\}$ with  cardinality  $|B_{i}|$, where $s$ is a solution (black dot) from the solution space $S$, and  $\mathcal{A}$ is the local search procedure.
A connection exists (blue dashed lines) between two local optima nodes if at least one solution in one basin has a neighbour solution in the other basin using the local search (neighbourhood) operator.

While many approaches to solving the TTP exist, only few rigorously analyse the difficulty of the problem based on the problem features.
\citet{wu2016impact} varied the renting rate parameter in an attempt to create hard-to-solve instances. 
\citet{yafrani2018fitness} used the local optima network representation to better understand the TTP's search space structure when using a local search heuristic joining the TSP and KP neighbourhoods~\cite{yafrani2017local}. The same representation is adopted by \cite{wuijts2019investigation}, who analysed fitness landscape characteristics from smaller instances of the problem, investigating the effectiveness of operators and metaheuristics that use local search.


In this paper, we extract LONs from multiple enumerable TTP instances using a neighbourhood search algorithm. This is done to investigate the impact of the problem features on the performance of local search by studying the topological structure of the problem search space.
The pseudocode of the local search is listed in Algorithm~\ref{algo:ls} where $F(.)$ is the TTP objective function, $\mathcal{N}_{TSP}(.)$ and $\mathcal{N}_{KP}(.)$ represent the neighbourhood functions for the TSP and KP components respectively. In the context of this study, we consider the local search named J2B~\cite{yafrani2017local} which uses the \emph{2-OPT} neighbourhood operator to generate $\mathcal{N}_{TSP}(.)$ and the one-bit-flip operator for the $\mathcal{N}_{KP}(.)$ neighbourhood.

{
\begin{algorithm}[htbp]
\small
\caption{Joint neighbourhood search algorithm}\label{algo:ls}
\begin{algorithmic}[1]
\STATE $s \gets $ initial solution
\WHILE{there is an improvement}
\FOR{each $s^* \in \mathcal{N}_{TSP}(s)$}
\FOR{each $s^{**} \in \mathcal{N}_{KP}(s*)$} 
\IF{$F(s^{**}) > F(s)$}
\STATE $s \gets s^{**}$
\ENDIF
\ENDFOR
\ENDFOR
\ENDWHILE
\end{algorithmic}
\end{algorithm}
}

While approaches with a sequential 
neighbourhood structure (iterating between the sub-problem neighbourhoods instead of combining them as in Algorithm~\ref{algo:ls})
report better results~\cite{mei2014improving, yafrani2016population}, they would result in a new LON for each sub-problem whenever the solution changes for the other sub-problem. Thus, the joint structure in Algorithm~\ref{algo:ls} was chosen due to this limitation.

\subsection{Interdependency models}
\label{sec:sect3}

\subsubsection{Standard model of interdependency}


The standard TTP model, $TTP_A$ (as per~\cite{polyakovskiy2014comprehensive}) is formulated as follows: given a set of $n$ cities, the associated matrix of distances $d_{ij}$, and a set of $m$ items distributed among these cities; each item $k$ is related with a profit $p_{k}$ and a weight $w_k$. The problem states that a thief should visit all the cities exactly once, stealing items on the road, returning to the starting city.
The knapsack capacity is denoted $W$,
the renting ratio, denoted $R$, dictates how much the thief should pay at the end of the travel with respect to the travel time, and $v_{max}$ and $v_{min}$ represent the maximum and minimum velocities respectively. Furthermore, each item is available in only one city, and $a_k \in \{1, \dots, n\}$ 
contains the reference of item $k$ to the corresponding city.

A TTP solution is represented by the tour $x = (x_1, \dots, x_n)$, a vector which contains the ordered list of cities; and the picking plan $z = (z_1, \dots, z_m)$, a binary vector coding the status of items ($1$ for ''packed'', and $0$ for ''not packed'').

The TTP was designed considering that the speed of the thief changes according to the knapsack weight, which makes the sub-problems mutually dependent. 
Therefore, the thief's velocity at city $x$ is defined in Equation~\ref{eq:velocity}.
\begin{equation}\label{eq:velocity}
v_x = v_{max} - K \cdot w^*_x
\end{equation}
where $K = ({v_{max}-v_{min}})/({W})$ is a constant value 
and $w^*_x$ is the weight of the knapsack at city $x$.

We note $g(z)= \sum_k p_k z_k, \quad \text{s.t.} \null  \sum_k w_k z_k \le W$ as the total items value and $f(x, z)= \sum_{i=1}^{n-1} t_{x_i, x_{i+1}} + t_{x_n, x_1} $ 
as the total travel time, where $t_{x_i, x_{i+1}} = {d_{x_i, x_{i+1}}}/{v_{x_i}}$ is the travel time from $x_i$ to $x_{i+1}$.

The objective is to maximise the total gain, $F_A(x, z) = g(z) - R \cdot f(x, z)$, by finding the best tour $x$ and picking plan $z$.


\subsubsection{Extended models of interdependency}
\label{sec:TTPext}

Here, we propose three additional models with different types of interdependencies. 

The first model, $TTP_0$, does not include any dependency, and it will serve for comparison with the other models. The objective function for $TTP_0$ is $F_0(x, z) = g(z) - R \cdot f(x)$.

The second TTP model, $TTP_B$, considers that the value of the items drops by time, and it does not consider the velocity-load dependency in Equation~\ref{eq:velocity}. Therefore the total value $g(x,z)$ depends on the tour too. 
The value of an item $k$ drops from $p_k$ to $p^{final}_k = p_k \cdot \mathcal{D}^{\ceil{\frac{T_k}{10}}}$, where $\mathcal{D} \in [0.1, 0.99]$ is the dropping rate, and $T_k$ is the carrying time.
To focus on the dependency analysis, we consider a linear combination of both objectives as $F_B(x, z) = g(x,z) - R \cdot f(x)$.

The last model, $TTP_C$, considers both types of dependency and its objective function is $F_C(x, z) = g(x,z) - R \cdot f(x,z)$.

The item value drop idea was introduced in~\cite{bonyadi2013travelling} to create a bi-objective version of the TTP ($TTP_2$). Here, we also use this idea in $TTP_B$ and $TTP_C$, however both are single-objective.

For the purpose of efficient implementations of neighbourhood-searching algorithms, the solution evaluation algorithms for $TTP_A$, $TTP_B$ and $TTP_C$ have the same worst-case complexity. However, the actual computation time for $TTP_A$ and $TTP_B$ can be significantly improved through caching when generating neighbouring solutions~\cite{yafrani2018efficiently,mei2014improving}. Such an approach is more difficult to achieve under the added item drop equation in $TTP_C$.

\section{Experimental results and discussion}
\label{sec:sect3}

Herein, we present the results of the empirical study of the the four models of interdependency. The instance are generated based on~\cite{bonyadi2013travelling, polyakovskiy2014comprehensive} and categorised based on the following features:

\begin{itemize}

\item \textbf{Profit-value correlation ($\mathcal{T}$): } This describes the correlation of the weight ($w_k$) and profit ($p_k$) of each item $k$. The TTP library addressed here considers three correlations, namely, \emph{uncorrelated (u)}, \emph{uncorrelated with similar weight (usw)}, and \emph{bounded strongly correlated (bsc)}.

\item \textbf{Knapsack capacity class ($\mathcal{C}$)}: This feature is a factor occurring in the maximum weight of the knapsack which is given by $W=\frac{\mathcal{C}}{11}\sum_{j=1}^m w_{j}$, where $\mathcal{C} \in \{2,5,10\}$ and $\frac{\mathcal{C}}{11}$ is applied as a limit for $W$, i.e., class $\mathcal{C}=10$ enlarges $W$ around to $90\%$, and more objects can be added in the knapsack~\cite{polyakovskiy2014comprehensive}.

\item \textbf{Item value dropping rate ($\mathcal{D}$): } This feature determines the rate at which the items lose value through time, and ranges between $0$ and $1$.
This feature is used to determine the value of an item $k$ at the end of the travel as follows $p_i=\mathcal{D}^{\lceil{\frac{T_k}{10}}\rceil}$, where $T_k$ is the total time that the item $k$ has been carried. 
\end{itemize}

We consider small instances to generate the local optima networks and identify the basins of attractions. Small instances have been chosen because the enumeration and study of the standard ones are impractical. The addressed instances contain $7$ cities and $6$ items (one per city, except for the starting one) and are generated following the guidelines by~\citet{polyakovskiy2014comprehensive}. 
The renting rate  is given by $R= ({g(Z_{opt})})/({f(X_{opt}, Z_{opt})})$,
where $Z_{opt}$ and $X_{opt}$ represent the optimal picking plan and the optimal tour respectively for the instances considered here.

\begin{table}[htbp]
\centering
\scriptsize
\caption{LON metrics for $TTP_0$ \label{tab:metrics_TTP0}}
\setlength\tabcolsep{3pt}
\begin{tabular}{cl|llllll} \hline
\multicolumn{1}{l}{$\mathcal{T}$} & \multicolumn{1}{l}{$\mathcal{C}$} & $\overline{n_v}$ & $\overline{n_e}$ & $\overline{C}$ & $\overline{C_r}$ & $\overline{l}$ & $\overline{|B|}$ \\ \hline

\multirow{3}{*}{\rotatebox{90}{$u$}}                & $2$                                & $1356_{875}$     & $19965_{10676}$  & $0.44_{0.17}$  & $0.05_{0.05}$   & $2.22_{0.24}$  & $6_{5}$        \\
                                    & $5$                                & $2678_{1752}$    & $72240_{42094}$  & $0.5_{0.13}$   & $0.04_{0.05}$   & $2.17_{0.19}$  & $12_{10}$      \\
                                    & $10$                               & $413_{379}$      & $16329_{18362}$  & $0.68_{0.11}$  & $0.34_{0.26}$   & $1.68_{0.27}$  & $274_{273}$    \\ \hline
\multirow{3}{*}{\rotatebox{90}{$usw$}}              & $2$                                & $4320_{0}$       & $52050_{0}$      & $0.31_{0}$     & $0.01_{0}$      & $2.4_{0}$      & $1_{0}$        \\
                                    & $5$                                & $10800_{0}$      & $260482_{0}$     & $0.26_{0}$     & $0_{0}$         & $2.58_{0}$     & $1_{0}$        \\
                                    & $10$                               & $3374_{255}$     & $157378_{11026}$ & $0.36_{0.01}$  & $0.03_{0}$      & $2.21_{0.03}$  & $14_{1}$       \\ \hline
\multirow{3}{*}{\rotatebox{90}{$bsc$}}              & $2$                                & $2040_{826}$     & $28539_{11355}$  & $0.32_{0.09}$  & $0.03_{0.04}$   & $2.32_{0.17}$  & $3_{3}$        \\
                                    & $5$                                & $3399_{3204}$    & $116929_{79643}$ & $0.4_{0.11}$   & $0.08_{0.09}$   & $2.19_{0.28}$  & $16_{18}$      \\
                                    & $10$                               & $781_{517}$      & $52839_{32613}$  & $0.65_{0.11}$  & $0.3_{0.23}$    & $1.72_{0.25}$  & $112_{110}$    \\ \hline

\end{tabular}
\end{table}

\begin{table}[htbp]
\centering
\scriptsize
\caption{LON metrics for $TTP_A$ \label{tab:metrics_TTPA}}
\setlength\tabcolsep{3pt}
\begin{tabular}{cl|llllll} \hline
\multicolumn{1}{l}{$\mathcal{T}$} & \multicolumn{1}{l}{$\mathcal{C}$} & $\overline{n_v}$ & $\overline{n_e}$ & $\overline{C}$ & $\overline{C_r}$ & $\overline{l}$ & $\overline{|B|}$ \\ \hline
\multirow{3}{*}{\rotatebox{90}{$u$}}   & $2$   & $483_{410}$    & $8324_{6074}$    & $0.6_{0.16}$   & $0.25_{0.3}$   & $1.89_{0.4}$   & $64_{147}$    \\
                       & $5$   & $105_{74}$     & $3013_{2496}$    & $0.85_{0.09}$  & $0.58_{0.25}$  & $1.42_{0.24}$  & $495_{635}$   \\
                       & $10$  & $25_{19}$      & $398_{530}$      & $0.97_{0.05}$  & $0.95_{0.09}$  & $1.05_{0.09}$  & $2777_{1950}$ \\ \hline
\multirow{3}{*}{\rotatebox{90}{$usw$}} & $2$   & $1643_{770}$   & $21453_{9261}$   & $0.5_{0.04}$   & $0.02_{0.01}$  & $2.31_{0.06}$  & $4_{2}$       \\
                       & $5$   & $3054_{1421}$  & $97404_{32996}$  & $0.57_{0.05}$  & $0.03_{0.03}$  & $2.15_{0.12}$  & $7_{4}$       \\
                       & $10$  & $125_{17}$     & $5559_{1379}$    & $0.84_{0.01}$  & $0.7_{0.01}$   & $1.3_{0.01}$   & $370_{62}$    \\ \hline
\multirow{3}{*}{\rotatebox{90}{$bsc$}} & $2$   & $1183_{814}$   & $17439_{10774}$  & $0.51_{0.14}$  & $0.12_{0.21}$  & $2.11_{0.34}$  & $13_{24}$     \\
                       & $5$   & $776_{1718}$   & $27057_{48137}$  & $0.76_{0.12}$  & $0.49_{0.28}$  & $1.55_{0.35}$  & $134_{119}$   \\
                       & $10$  & $58_{21}$      & $1537_{1027}$    & $0.92_{0.04}$  & $0.87_{0.07}$  & $1.13_{0.07}$  & $836_{276}$   \\ \hline

\end{tabular}
\end{table}

\begin{table}[htbp]
\centering
\scriptsize
\caption{LON metrics for $TTP_B$ \label{tab:metrics_TTPB}}
\setlength\tabcolsep{2pt}
\begin{tabular}{cll|lllllll}\hline
\multicolumn{1}{l}{$\mathcal{T}$} & \multicolumn{1}{l}{$\mathcal{C}$} & \multicolumn{1}{l}{$\mathcal{D}$} & $\overline{n_v}$ & $\overline{n_e}$ & $\overline{C}$ & $\overline{C_r}$ & $\overline{l}$ & $\overline{|B|}$ \\ \hline

\multirow{9}{*}{\rotatebox{90}{$u$}}   & \multirow{3}{*}{$2$}  & $0.9$  & $741_{634}$   & $10432_{8838}$   & $0.57_{0.21}$  & $0.17_{0.25}$ & $2.02_{0.39}$ & $33_{44}$   \\ 
                       &                       & $0.95$ & $1263_{823}$  & $18732_{9734}$   & $0.5_{0.18}$   & $0.05_{0.05}$ & $2.16_{0.18}$ & $7_{10}$    \\ 
                       &                       & $0.98$ & $626_{513}$   & $9788_{6313}$    & $0.53_{0.18}$  & $0.17_{0.24}$ & $1.99_{0.35}$ & $24_{40}$   \\ \cline{2-9} 
                       & \multirow{3}{*}{$5$}  & $0.9$  & $303_{192}$   & $12087_{9042}$   & $0.76_{0.06}$  & $0.34_{0.2}$  & $1.66_{0.21}$ & $104_{80}$  \\ 
                       &                       & $0.95$ & $944_{773}$   & $34083_{29961}$  & $0.64_{0.12}$  & $0.21_{0.25}$ & $1.85_{0.29}$ & $67_{80}$   \\ 
                       &                       & $0.98$ & $1247_{813}$  & $41802_{26987}$  & $0.61_{0.07}$  & $0.08_{0.06}$ & $1.99_{0.11}$ & $27_{21}$   \\ \cline{2-9} 
                       & \multirow{3}{*}{$10$} & $0.9$  & $67_{28}$     & $1955_{1387}$    & $0.9_{0.05}$   & $0.82_{0.1}$  & $1.18_{0.1}$  & $777_{337}$ \\ 
                       &                       & $0.95$ & $105_{57}$    & $4072_{3690}$    & $0.83_{0.05}$  & $0.69_{0.13}$ & $1.31_{0.13}$ & $517_{203}$ \\ 
                       &                       & $0.98$ & $380_{412}$   & $19659_{24222}$  & $0.75_{0.16}$  & $0.46_{0.33}$ & $1.54_{0.33}$ & $419_{419}$ \\ \hline
\multirow{9}{*}{\rotatebox{90}{$usw$}} & \multirow{3}{*}{$2$}  & $0.9$  & $3761_{443}$  & $46009_{5229}$   & $0.45_{0.06}$  & $0.01_{0}$    & $2.36_{0.02}$ & $1_{0}$     \\ 
                       &                       & $0.95$ & $4283_{55}$   & $51701_{537}$    & $0.33_{0.02}$  & $0.01_{0}$    & $2.39_{0.01}$ & $1_{0}$     \\ 
                       &                       & $0.98$ & $4320_{0}$    & $52049_{69}$     & $0.31_{0}$     & $0.01_{0}$    & $2.4_{0}$     & $1_{0}$     \\ \cline{2-9} 
                       & \multirow{3}{*}{$5$}  & $0.9$  & $2789_{391}$  & $91542_{8516}$   & $0.6_{0.01}$   & $0.02_{0}$    & $2.16_{0.04}$ & $6_{1}$     \\ 
                       &                       & $0.95$ & $6470_{467}$  & $172352_{10217}$ & $0.47_{0.01}$  & $0.01_{0}$    & $2.3_{0.01}$  & $2_{0}$     \\ 
                       &                       & $0.98$ & $9947_{708}$  & $245193_{14408}$ & $0.33_{0.03}$  & $0_{0}$       & $2.43_{0.02}$ & $2_{0}$     \\ \cline{2-9} 
                       & \multirow{3}{*}{$10$} & $0.9$  & $167_{5}$     & $8527_{300}$     & $0.81_{0}$     & $0.62_{0.01}$ & $1.38_{0.01}$ & $272_{8}$   \\ 
                       &                       & $0.95$ & $514_{75}$    & $34247_{5195}$   & $0.69_{0.01}$  & $0.26_{0.04}$ & $1.74_{0.04}$ & $90_{13}$   \\ 
                       &                       & $0.98$ & $1473_{157}$  & $84940_{7299}$   & $0.58_{0.01}$  & $0.08_{0.01}$ & $1.95_{0.02}$ & $31_{3}$    \\ \hline
\multirow{9}{*}{\rotatebox{90}{$bsc$}} & \multirow{3}{*}{$2$}  & $0.9$  & $1367_{985}$  & $18580_{10718}$  & $0.53_{0.14}$  & $0.15_{0.25}$ & $2.09_{0.4}$  & $15_{28}$   \\ 
                       &                       & $0.95$ & $1003_{846}$  & $19637_{12444}$  & $0.49_{0.14}$  & $0.1_{0.09}$  & $2.05_{0.25}$ & $12_{11}$   \\ 
                       &                       & $0.98$ & $1978_{932}$  & $29963_{13027}$  & $0.33_{0.14}$  & $0.05_{0.1}$  & $2.34_{0.26}$ & $8_{17}$    \\ \cline{2-9} 
                       & \multirow{3}{*}{$5$}  & $0.9$  & $560_{973}$   & $20766_{31803}$  & $0.77_{0.13}$  & $0.45_{0.34}$ & $1.57_{0.38}$ & $175_{185}$ \\ 
                       &                       & $0.95$ & $1262_{597}$  & $64100_{30502}$  & $0.56_{0.09}$  & $0.13_{0.13}$ & $1.96_{0.2}$  & $24_{25}$   \\ 
                       &                       & $0.98$ & $3351_{2567}$ & $124014_{89315}$ & $0.46_{0.13}$  & $0.08_{0.14}$ & $2.17_{0.29}$ & $18_{28}$   \\ \cline{2-9} 
                       & \multirow{3}{*}{$10$} & $0.9$  & $85_{37}$     & $3351_{1979}$    & $0.91_{0.05}$  & $0.85_{0.09}$ & $1.15_{0.09}$ & $725_{644}$ \\ 
                       &                       & $0.95$ & $212_{109}$   & $13849_{8907}$   & $0.82_{0.07}$  & $0.65_{0.19}$ & $1.35_{0.19}$ & $283_{185}$ \\ 
                       &                       & $0.98$ & $462_{444}$   & $28893_{25806}$  & $0.75_{0.14}$  & $0.52_{0.31}$ & $1.49_{0.32}$ & $194_{136}$ \\ \hline

\end{tabular}
\end{table}

\begin{table}[htbp]
\centering
\scriptsize
\caption{LON metrics for $TTP_C$ \label{tab:metrics_TTPC}}
\setlength\tabcolsep{2pt}
\begin{tabular}{cll|lllllll}\hline
\multicolumn{1}{l}{$\mathcal{T}$} & \multicolumn{1}{l}{$\mathcal{C}$} & \multicolumn{1}{l}{$\mathcal{D}$} & $\overline{n_v}$ & $\overline{n_e}$ & $\overline{C}$ & $\overline{C_r}$ & $\overline{l}$ & $\overline{|B|}$ \\ \hline

\multirow{9}{*}{\rotatebox{90}{$u$}}   & \multirow{3}{*}{$2$}  & $0.9$  & $276_{215}$  & $5728_{4859}$   & $0.58_{0.12}$ & $0.25_{0.17}$ & $1.82_{0.24}$ & $28_{29}$     \\
                       &                       & $0.95$ & $351_{234}$  & $5826_{4307}$   & $0.65_{0.17}$ & $0.19_{0.19}$ & $1.9_{0.27}$  & $24_{23}$     \\
                       &                       & $0.98$ & $191_{300}$  & $3952_{7226}$   & $0.76_{0.18}$ & $0.54_{0.3}$  & $1.5_{0.38}$  & $105_{103}$   \\ \cline{2-9}
                       & \multirow{3}{*}{$5$}  & $0.9$  & $40_{18}$    & $777_{582}$     & $0.93_{0.04}$ & $0.88_{0.09}$ & $1.12_{0.09}$ & $658_{390}$   \\
                       &                       & $0.95$ & $63_{73}$    & $2017_{3582}$   & $0.91_{0.08}$ & $0.82_{0.19}$ & $1.17_{0.19}$ & $681_{552}$   \\
                       &                       & $0.98$ & $71_{49}$    & $2072_{1875}$   & $0.86_{0.07}$ & $0.69_{0.17}$ & $1.31_{0.17}$ & $632_{636}$   \\ \cline{2-9}
                       & \multirow{3}{*}{$10$} & $0.9$  & $20_{8}$     & $216_{166}$     & $0.99_{0.01}$ & $0.99_{0.02}$ & $1.01_{0.02}$ & $2538_{1150}$ \\
                       &                       & $0.95$ & $19_{8}$     & $200_{157}$     & $0.99_{0.02}$ & $0.98_{0.02}$ & $1.02_{0.02}$ & $2849_{1413}$ \\
                       &                       & $0.98$ & $27_{13}$    & $393_{350}$     & $0.97_{0.03}$ & $0.96_{0.04}$ & $1.04_{0.04}$ & $2032_{949}$  \\ \hline
\multirow{9}{*}{\rotatebox{90}{$usw$}} & \multirow{3}{*}{$2$}  & $0.9$  & $730_{93}$   & $10646_{1026}$  & $0.41_{0.03}$ & $0.04_{0.01}$ & $2.25_{0.03}$ & $7_{1}$       \\
                       &                       & $0.95$ & $816_{166}$  & $11619_{1909}$  & $0.44_{0.04}$ & $0.04_{0.01}$ & $2.26_{0.03}$ & $6_{1}$       \\
                       &                       & $0.98$ & $1060_{291}$ & $14504_{3482}$  & $0.47_{0.03}$ & $0.03_{0.01}$ & $2.29_{0.04}$ & $5_{2}$       \\ \cline{2-9}
                       & \multirow{3}{*}{$5$}  & $0.9$  & $180_{77}$   & $9730_{5706}$   & $0.79_{0.06}$ & $0.62_{0.15}$ & $1.38_{0.15}$ & $107_{49}$    \\
                       &                       & $0.95$ & $408_{102}$  & $26383_{7503}$  & $0.71_{0.03}$ & $0.33_{0.07}$ & $1.67_{0.07}$ & $41_{11}$     \\
                       &                       & $0.98$ & $739_{319}$  & $38967_{13790}$ & $0.67_{0.04}$ & $0.18_{0.1}$  & $1.84_{0.12}$ & $26_{13}$     \\ \cline{2-9}
                       & \multirow{3}{*}{$10$} & $0.9$  & $25_{7}$     & $305_{190}$     & $0.99_{0.01}$ & $0.99_{0.02}$ & $1.01_{0.02}$ & $1960_{424}$  \\
                       &                       & $0.95$ & $52_{3}$     & $1255_{165}$    & $0.95_{0.01}$ & $0.94_{0.02}$ & $1.06_{0.02}$ & $872_{58}$    \\
                       &                       & $0.98$ & $42_{11}$    & $886_{499}$     & $0.97_{0.01}$ & $0.96_{0.01}$ & $1.04_{0.01}$ & $1137_{256}$  \\ \hline
\multirow{9}{*}{\rotatebox{90}{$bsc$}} & \multirow{3}{*}{$2$}  & $0.9$  & $426_{286}$  & $6819_{3726}$   & $0.59_{0.19}$ & $0.24_{0.34}$ & $1.89_{0.43}$ & $36_{65}$     \\
                       &                       & $0.95$ & $232_{219}$  & $4927_{5598}$   & $0.69_{0.14}$ & $0.36_{0.27}$ & $1.69_{0.33}$ & $51_{44}$     \\
                       &                       & $0.98$ & $505_{384}$  & $9119_{7403}$   & $0.59_{0.15}$ & $0.22_{0.25}$ & $1.9_{0.36}$  & $31_{40}$     \\ \cline{2-9}
                       & \multirow{3}{*}{$5$}  & $0.9$  & $61_{87}$    & $2539_{6173}$   & $0.93_{0.1}$  & $0.88_{0.19}$ & $1.12_{0.19}$ & $697_{408}$   \\
                       &                       & $0.95$ & $79_{34}$    & $2121_{1170}$   & $0.87_{0.08}$ & $0.75_{0.22}$ & $1.25_{0.22}$ & $298_{142}$   \\
                       &                       & $0.98$ & $211_{187}$  & $9749_{10470}$  & $0.81_{0.11}$ & $0.56_{0.31}$ & $1.44_{0.31}$ & $252_{230}$   \\ \cline{2-9}
                       & \multirow{3}{*}{$10$} & $0.9$  & $26_{8}$     & $338_{214}$     & $0.98_{0.02}$ & $0.98_{0.02}$ & $1.02_{0.02}$ & $1834_{522}$  \\
                       &                       & $0.95$ & $41_{12}$    & $803_{418}$     & $0.96_{0.03}$ & $0.94_{0.05}$ & $1.06_{0.05}$ & $1197_{431}$  \\
                       &                       & $0.98$ & $48_{21}$    & $1205_{929}$    & $0.95_{0.03}$ & $0.93_{0.05}$ & $1.07_{0.05}$ & $1105_{535}$  \\ \hline

\end{tabular}
\end{table}

\begin{figure}[htbp]
\centering
\hspace{-3mm}
\subfloat[$\mathcal{T}=u$]{
\includegraphics[width=0.155\textwidth,trim=0 0 0 33, clip]{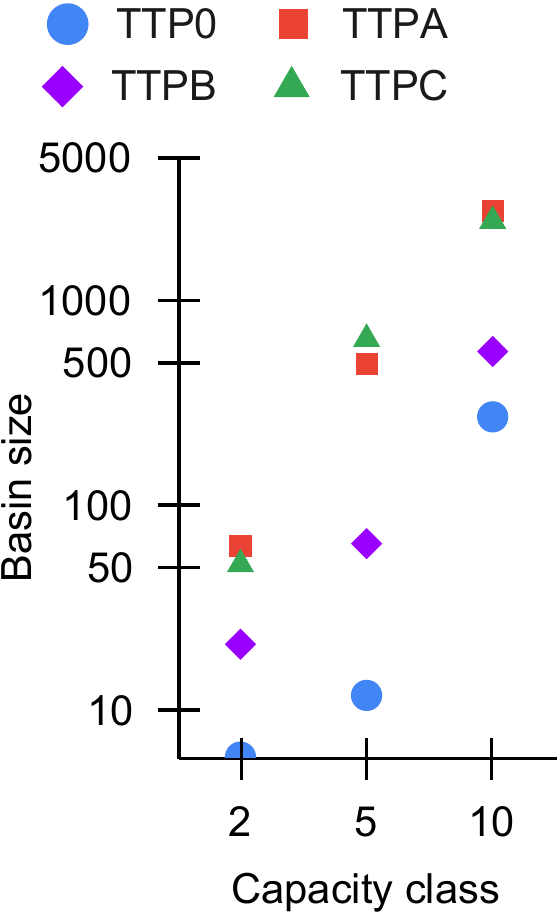}
}\hspace{-6mm}
\quad
\subfloat[$\mathcal{T}=usw$]{
\includegraphics[width=0.155\textwidth]{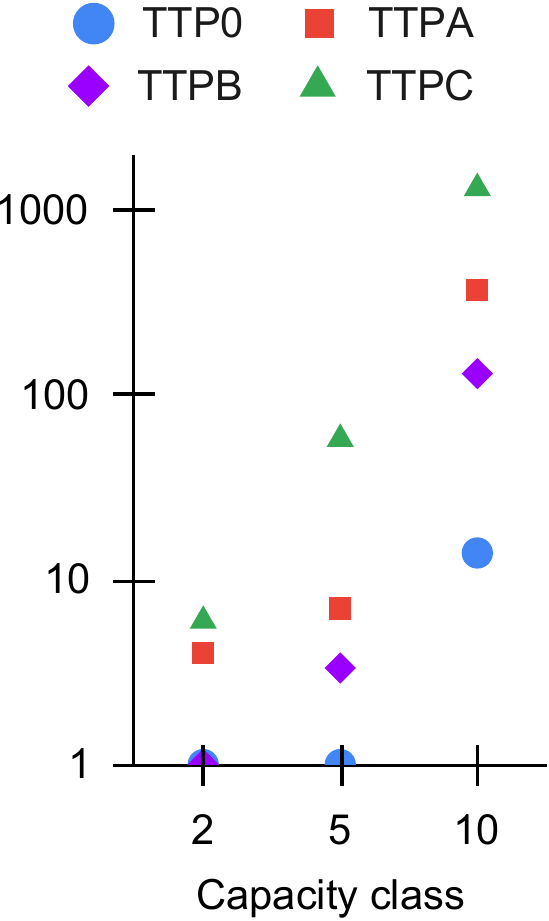}
}\hspace{-6mm}
\quad
\subfloat[$\mathcal{T}=bsc$]{
\includegraphics[width=0.155\textwidth,trim=0 0 0 33, clip]{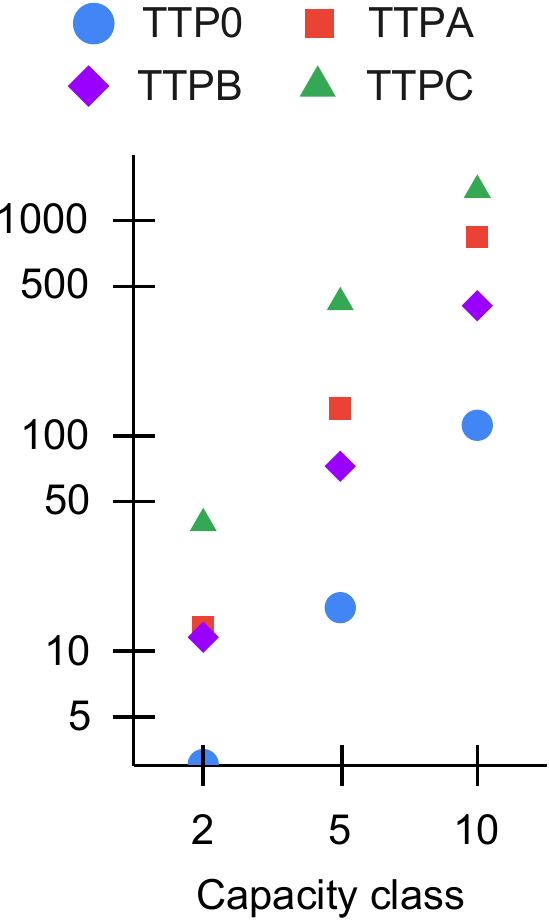}
}
\hspace{-4mm}
\caption{Comparison of the average basin sizes. For $TTP_B$ and $TTP_C$ the averages across the three $\mathcal{D}$ values is used}
\label{fig:standard-extended}
\end{figure}

During the instance generation process, the TSP component is fixed, i.e., the set of coordinates is the same for all the instances. We use three capacity classes $\mathcal{C} \in \{2,5,10\}$, three dropping rates $\mathcal{D} \in \{0.9, 0.95, 0.98\}$ (for $TTP_B$ and $TTP_C$), and  all three correlation variants ($u$, $usw$, and $bsc$).
This results in $9$ classes for $TTP_0$ and $TTP_A$, and in $27$ classes for $TTP_B$ and $TTP_C$. For each class, $100$ instances are generated, and the corresponding LONs are extracted to analyse their fitness landscapes. 

Tables~\ref{tab:metrics_TTP0}, \ref{tab:metrics_TTPA}, \ref{tab:metrics_TTPB} and \ref{tab:metrics_TTPC} show the graph metrics for the standard and extended models, with the subscript numbers representing standard deviations.
The results in Tables~\ref{tab:metrics_TTPA} and~\ref{tab:metrics_TTPC} are aligned with the findings in~\cite{yafrani2018fitness} and can be summarised as follows.
When the knapsack capacity is increased (higher $\mathcal{C}$), the number of nodes ($\overline{n_v}$) and edges ($\overline{n_e}$) decreases, while the basin size ($\overline{|B|}$) increases. This means that the landscape is easy to navigate when the knapsack capacity is large. This trend is sometimes broken when the values of items are uncorrelated with the weights, and the weights are similar ($\mathcal{T}=usw$).

The average path lengths ($\overline{l}$) are always small, which shows that the transition from a random local optimum to another is done through very few intermediate local optima. This fact, combined with the observation that the clustering coefficient ($\overline{C}$) is always higher than the clustering coefficient of the equivalent Erdős–Rényi random graph ($\overline{C_r}$), 
indicates that the LONs have small-world properties \cite{humphries2008network}. It is worth mentioning that when the LON has a small number of nodes, the corresponding random graph naturally has a high clustering coefficient ($\overline{C_r}$), which can be explained by the small number of combinations to connect the nodes.

In Tables~\ref{tab:metrics_TTPB} and~\ref{tab:metrics_TTPC}, it is difficult to isolate the impact of changing the dropping rate ($\mathcal{D}$). In Table~\ref{tab:metrics_TTPC}, the most prominent trend is a positive correlation between $\mathcal{D}$ and the basin size $\overline{|B|}$. However, more samples over $\mathcal{D}$ are required to isolate the outliers if any and understand its impact.


Interestingly, in Figure~\ref{fig:standard-extended}, when comparing the basin sizes of all models, we observe that the basin sizes of $TTP_0$ are mostly smaller than the remaining models. This is suggesting that the models with dependencies result in landscapes that are easier to navigate compared to the dependency-free model $TTP_0$.

\begin{figure}[htbp]
\subfloat[$TTP_0$]{
\includegraphics[width=0.24\textwidth,trim=0 20 0 0,clip]{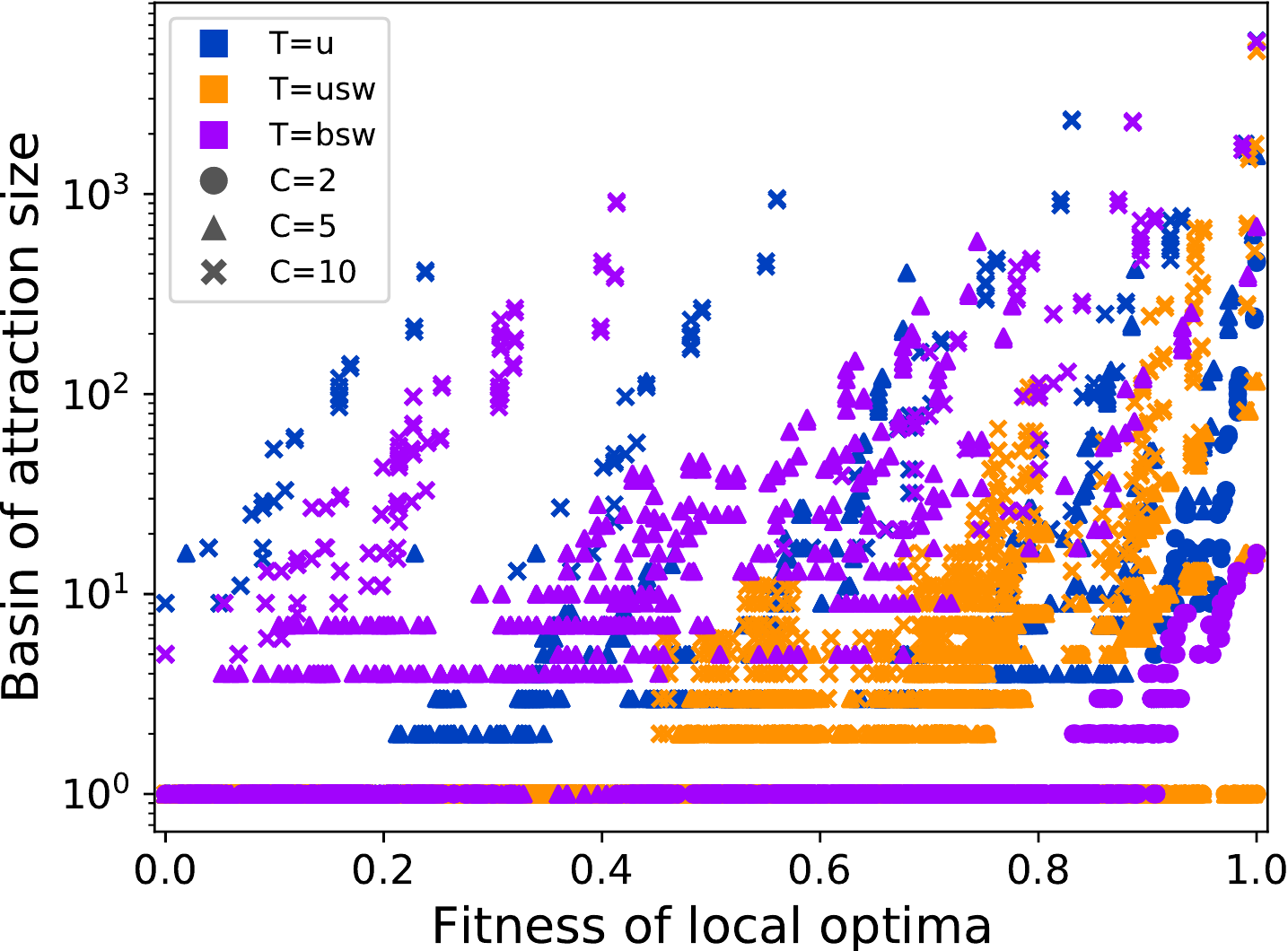}
}
\subfloat[$TTP_A$]{
\includegraphics[width=0.231\textwidth,trim=20 20 0 0,clip]{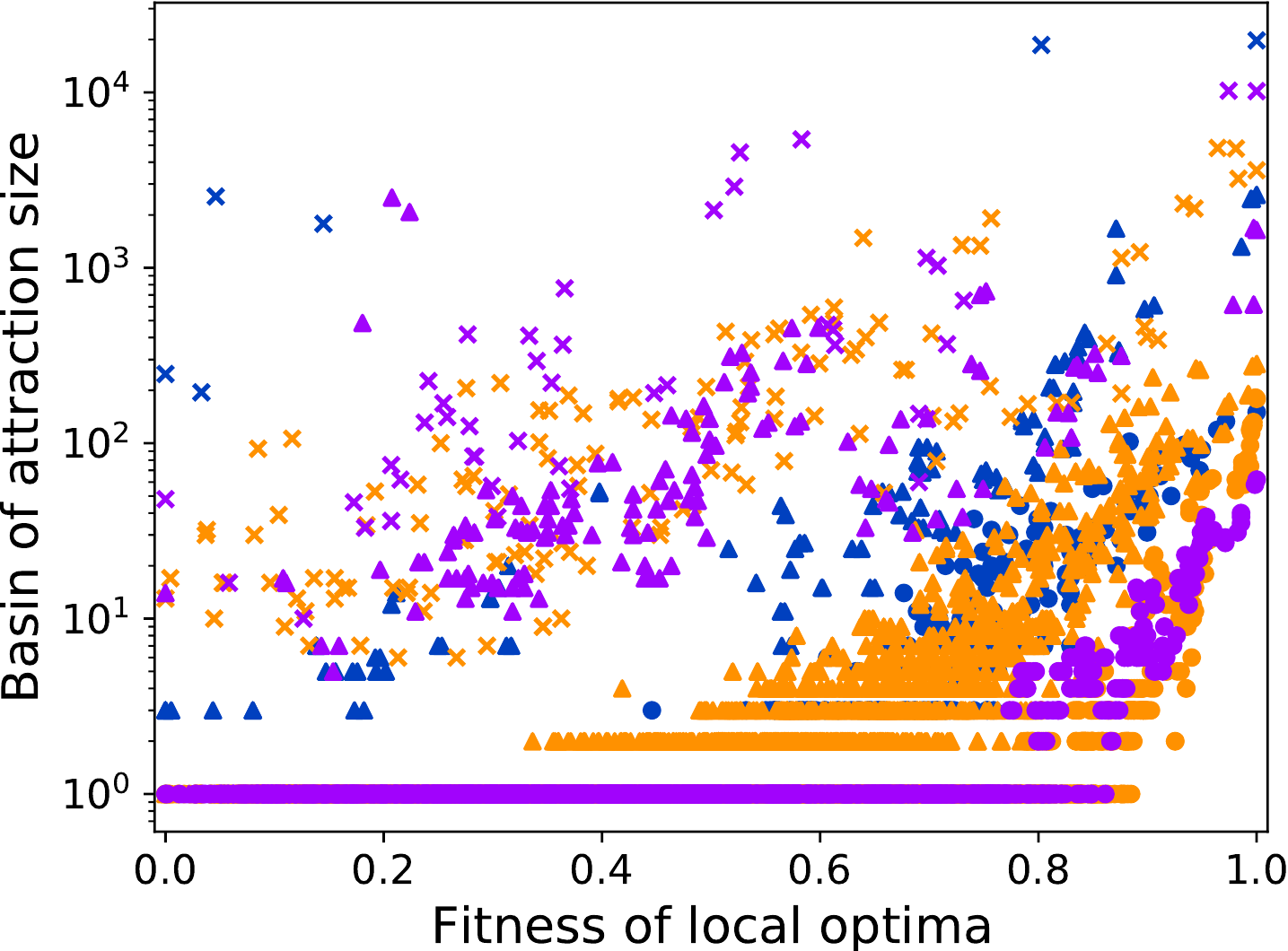}
}
\vspace{-3mm}
\subfloat[$TTP_B$]{
\includegraphics[width=0.24\textwidth]{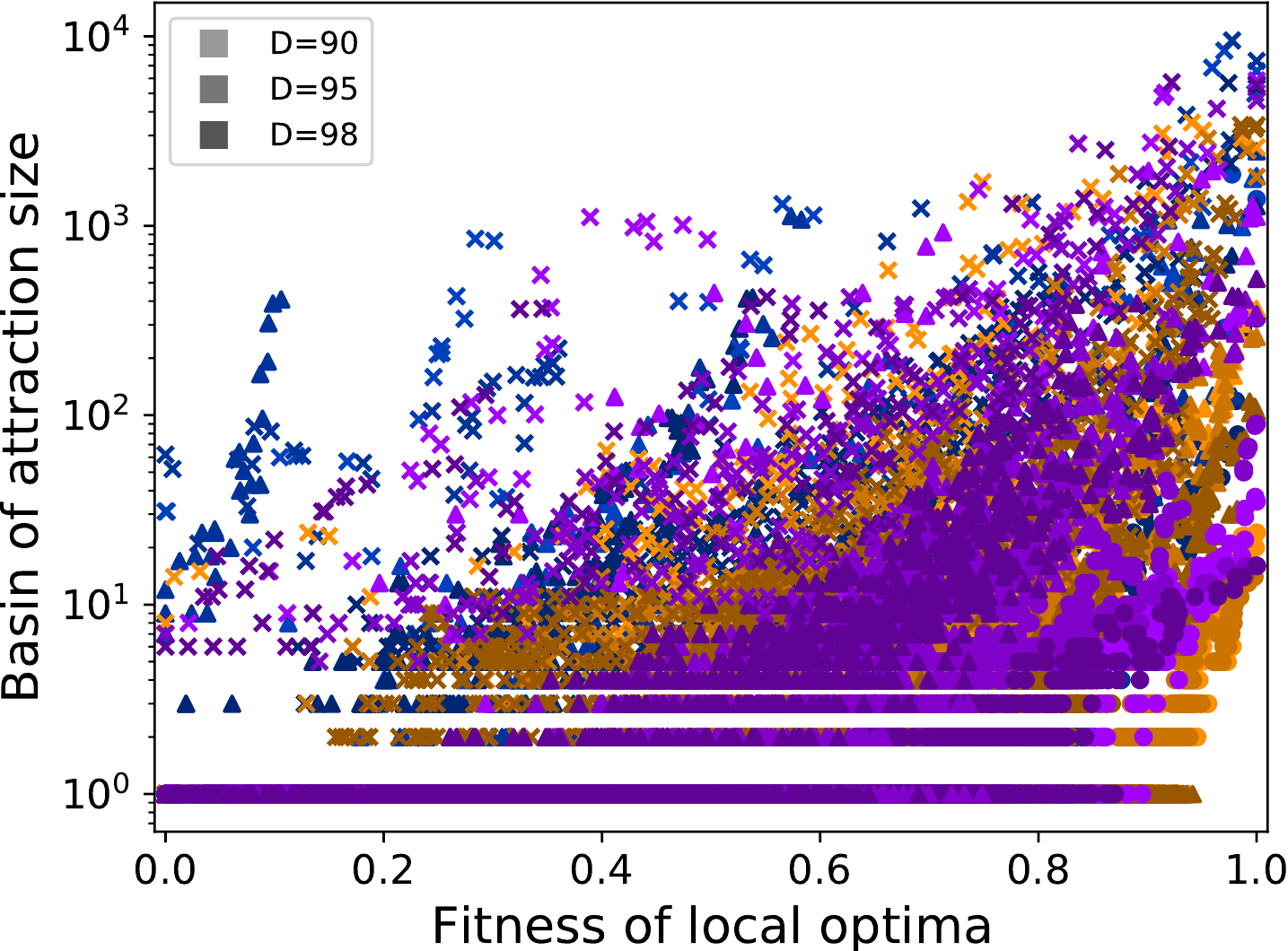}
}
\subfloat[$TTP_C$]{
\includegraphics[width=0.231\textwidth,trim=20 0 0 0,clip]{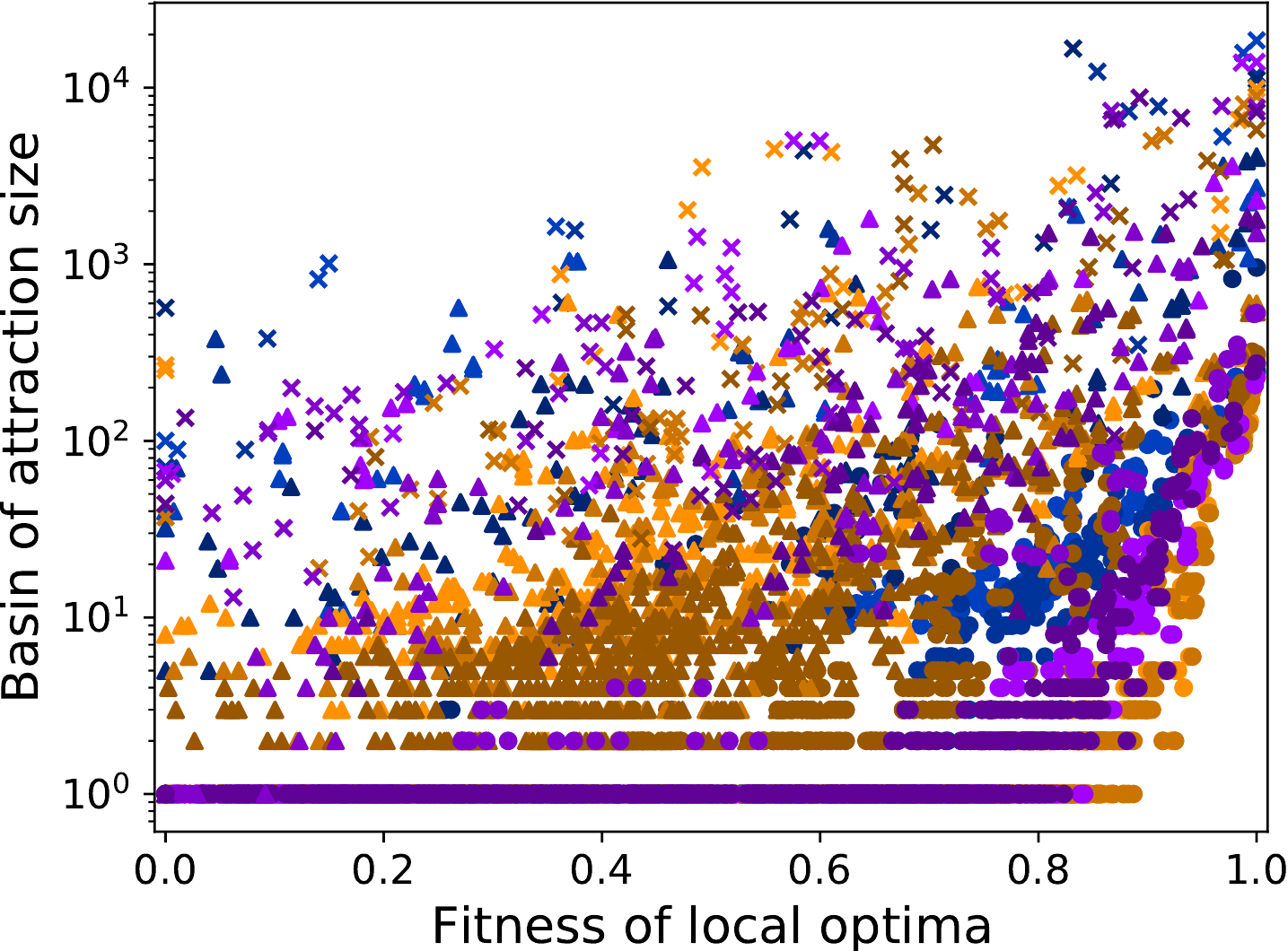}
}
\vspace{-2mm}
\caption{Fitness-basin size correlations. Shown for one random instance per category.}
\label{fig:scatter-fb}
\end{figure}

To further investigate the difficulty of the different models, we consider Spearman's correlation coefficients between the fitness and basin size for all models, where the average is calculated using Fisher z-transformation~\cite{fieller1961tests}. The correlations are as follows: $\rho(TTP_0)=0.61$, $\rho(TTP_A)=0.79$, $\rho(TTP_B)=0.74$, $\rho(TTP_C)=0.80$, and the associated scatter plots are shown in Figure~\ref{fig:scatter-fb}.
There is a positive correlation between the fitnesses and basin sizes for all $4$ models, showing that larger basins tend to have a higher fitness. The correlation is clearly the weakest for $TTP_0$, i.e., it appears to be the most difficult there to find an optimal solution. This correlation can be seen for individual categories in Figure~\ref{fig:scatter-fb}.

In summary, it is surprising that the introduction of interdependencies appears to make the problem easier for the given algorithm. We can only speculate about the reasons. Firstly, it might be that the joint neighbourhood search induces unnecessary complexity to the $TTP_0$ landscapes. Indeed, there is no interdependency in $TTP_0$, i.e., the instances can simply be solved by solving the sub-problems separately. Secondly, maybe our analyses are inadequate, and new powerful tools are needed for characterising landscapes.

\section{Conclusion}
\label{sec:sect5}



In this paper we have investigated the impact of different forms of dependency on the difficulty of solving problems with multiple interconnected sub-problems using local search. We considered the Travelling Thief Problem (TTP) as our study case and investigated three extended models embedding other forms of dependency, and compared it to the standard TTP model by analysing how dependency impacts the solution landscape of the problem. 

The analysis was conducted on enumerable instances using Local Optima Networks (LONs) and topology metrics for a local search algorithm which combines the TSP and KP neighbourhoods. 
The preliminary results gave us some insights on the impact of the dependency equations 
to the standard TTP. Specifically, the addition does not appear to result in more difficult search landscapes for the considered algorithm.

Nevertheless, this study has some limitations and aspects that should be further investigated. Firstly, joint neighbourhood search algorithms are not the most efficient in practice due to their high computational complexity. Thus, the analysis does not necessarily generalise to other types of local search with a sequential neighbourhood structure. Along similar lines, our results might not carry over to non-enumerable instances; however, if we investigate larger instances, the characteristics of the sampled LONs can be highly susceptible to the sampling approach~\cite{jakobovic2021boolean}. 
Secondly, the item drop feature ($\mathcal{D}$) should be further investigated to better isolate its impact. Lastly, to overcome the limitations induced by LONs and composite structure of the TTP, regression (cost) models may be able provide additional insights on the behaviour of (meta-)heuristics on the TTP models.



\bibliographystyle{ACM-Reference-Format}
\bibliography{references}

\end{document}